\documentclass[runningheads]{llncs}
\usepackage{graphicx}
\usepackage[T1]{fontenc}

\usepackage{tikz}
\usepackage{comment}
\usepackage{amsmath,amssymb} 
\usepackage{color}
\usepackage{orcidlink}

\usepackage[accsupp]{axessibility}  

\begin{document}
\pagestyle{headings}
\mainmatter
\def\ECCVSubNumber{192}  

\title{Towards Better Guided Attention and Human Knowledge Insertion in Deep Convolutional Neural Networks} 

\titlerunning{Multi-Scale Attention Branch Networks}
%
\author{Ankit Gupta\inst{1}\orcidlink{0000-0002-9961-1041}\index{Gupta, Ankit} \and
Ida-Maria Sintorn\inst{1,2}\orcidlink{0000-0002-8307-7411} 
}
\authorrunning{A. Gupta, I. Sintorn}
%
\institute{Department of Information Technology, Uppsala
University, 75236 Uppsala, Sweden \and
Vironova AB, 11330 Gävlegatan 22, Stockholm, Sweden
\email{\{ankit.gupta,ida.sintorn\}@it.uu.se}
}

\maketitle

\begin{abstract}
Attention Branch Networks (ABNs) have been shown to simultaneously provide visual explanation and improve the performance of deep convolutional neural networks (CNNs). In this work, we introduce Multi-Scale Attention Branch Networks (MSABN), which enhance the resolution of the generated attention maps, and improve the performance. We evaluate MSABN on benchmark image recognition and fine-grained recognition datasets where we observe MSABN outperforms ABN and baseline models. We also introduce a new data augmentation strategy utilizing the attention maps to incorporate human knowledge in the form of bounding box annotations of the objects of interest. We show that even with a limited number of edited samples, a significant performance gain can be achieved with this strategy.

\end{abstract}

\section{Introduction}
CNNs have established themselves as the benchmark approach in image recognition \cite{he2016deep,krizhevsky2012imagenet,xie2017aggregated,tan2019efficientnet}. However, the interpretation of the decision-making process still remains elusive. To be able to visualize and verify that the decision of a CNN is based on correct and meaningful information is important for many computer vision applications like self-driving cars or automated social media content analysis, to increase reliability and trust in the system. For biomedical applications, where instrument settings and small cohorts/datasets easily bias the results this becomes extremely important. This, in addition to the high cost or risk associated with erroneous decisions, limit the reliability and hence also the deployment of CNN-based solutions in clinical diagnostics and biomedical analysis. Tools to explain the decision and be able to correct erroneous conclusions drawn by CNN would improve the trustworthiness of the technology.

Visual explanation \cite{zhang2018visual} is used in deep learning to interpret the decisions of the CNNs. Broadly, these methods can be categorized as either requiring additional backpropagation or not. Methods requiring backpropagation \cite{selvaraju2017grad,chattopadhay2018grad} can be used out-of-the-box and don't require any network architectural or training method changes. However, they need an extra backpropagation step to find the discriminative regions in images. Response-based methods \cite{zhou2016learning,fukui2019attention} don't require backpropagation because they generate explanations and predictions simultaneously. In this work, we improve upon \cite{fukui2019attention} and present a response-based visual explanation method that outperforms previous methods.

To fully utilize the benefits of deep learning in practical settings, it would also be beneficial to be able to incorporate human knowledge and interact with the attention maps to correct and improve the networks. One attempt to do this is presented in \cite{mitsuhara2019embedding}, where the attention map of the images is manually edited and then the network is fine-tuned with this human knowledge. This approach, however, requires manual detailed annotation (drawing) of the attention maps which becomes tedious for large datasets. Here, we propose a softer form of user input in the form of object bounding boxes. 

\begin{figure}        
    \centering
    \includegraphics[width=0.8\columnwidth]{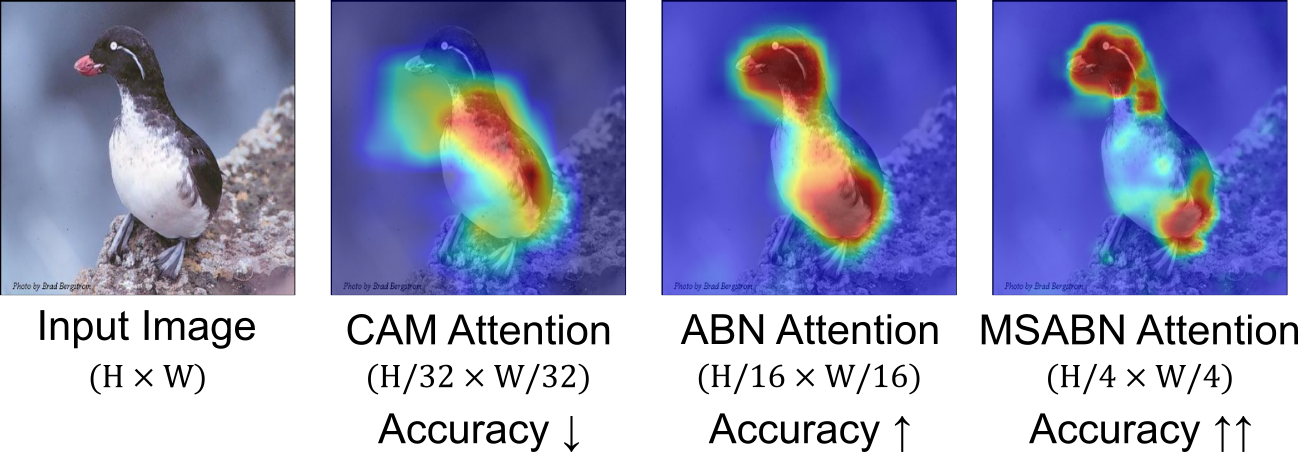}
    \caption{Comparison of the MSABN attention map with CAM and ABN attention outputs. MSABN provides better accuracy and higher resolution attention maps. }
    \label{fig:intro_fig}
\end{figure}

In this paper, we build upon the attention branch network structure and present ways to improve the attention maps and overall network performance even further. In addition, we present a simple and efficient way to incorporate human expertise when training/refining the networks. Figure \ref{fig:intro_fig} illustrates the difference in the attention map detail achieved with our proposed multi-scale attention branch network (MSABN) compared to the attention branch network (ABN) and the commonly used Class Activation Mapping (CAM). 

Our main contributions are:
\begin{itemize}
    \item With the introduction of MSABN, we increase the performance of the response-based guided attention models significantly and simultaneously increase the resolution of attention maps by 4x which has not been achieved with the response-based methods yet.
    \item With the introduction of the puzzle module in the attention branch, the performance is improved further and it also adds better localization performance for fine-grained recognition.
    \item We provide a human-in-the-loop (HITL) pipeline for inserting human knowledge in the models which require simple annotation (object bounding box) and achieve performance improvement even by annotating/correcting a small portion of the dataset.
\end{itemize}

\section{Related Work} 

\subsubsection{Visual explanation:} Highlighting the regions or details of an image that are important for the decision is commonly used to interpret and verify the decision-making process of CNNs \cite{selvaraju2017grad,bach2015pixel}. Overall, the methods for visual explanation fall into two categories: gradient-based and response-based. Gradient-based methods (SmoothGrad \cite{smilkov2017smoothgrad}, Guided Backpropagation \cite{springenberg2014striving}, LIME \cite{ribeiro2016should}, grad-CAM \cite{selvaraju2017grad}, gradCAM++ \cite{chattopadhay2018grad}, LRP \cite{bach2015pixel}) rely on the backpropagation of auxiliary data like class index or noise to find the discriminative regions in the image. These methods do not require training or architectural modifications. However, backpropagation adds another step during inference and increases the computational cost. Response-based methods, on the other hand, do not require backpropagation and can provide interpretation at the same time as inference. CAM \cite{zhou2016learning}, is a response-based method that provides an attention map for each class by replacing the final fully-connected layer of a CNN with a convolutional layer and global average pooling. The recently proposed attention branch networks \cite{fukui2019attention} extend the response-based model by adding a branch structure to provide the attention to the network in a feed-forward way and show that adding attention in this way improves the performance of the models.

\subsubsection{Inserting human knowledge:}
Human-in-the-loop (HITL) refers to approaches in the deep learning framework where a prediction model is trained with integrated human knowledge or user guidance. Several HITL approaches \cite{branson2010visual,branson2011strong,wang2016human,parkash2012attributes,parikh2011interactively} have been used to inject human knowledge in different ways into the deep neural networks. In \cite{linsley2019learning}, the ClickMe interface is used to get human annotation in the form of mouse clicks to get approximate attention regions in an image. They also modified the model architecture to incorporate this human knowledge and demonstrated improved performance. In \cite{rieger2020interpretations}, the authors proposed a method that leverages the explanation generated by the models such that they should provide the correct explanation for the correct prediction. In \cite{mitsuhara2019embedding}, Mitsuhara, M. et al. showed that editing the attention maps and sequentially fine-tuning the ABN models improve performance as mentioned in the Introduction. However, their approach involves manually marking the object boundary in a large number (thousands) of images making it unfeasible in practice. In \cite{kc2021improving,arai2021non}, the authors attempted to improve model performance by using non-strict attention input and introducing different loss functions and managing to improve performance. In \cite{kc2021improving}, Dharma et al. use different loss functions for the regions inside and outside the object bounding box as a way to incorporate human input. Our method is similar to these methods but focused on augmenting the data rather than changing the loss function.

\subsubsection{Copy-paste augmentation:} This refers to an augmentation strategy where a patch from one image is pasted into another image which is then used for training. In 
\cite{dvornik2018modeling,dwibedi2017cut,ghiasi2021simple} this augmentation strategy is used to improve the performance in instance segmentation and detection tasks. The idea is that by cutting object instances and pasting them into other images a different context for the objects is provided. The CutMix \cite{yun2019cutmix} augmentation strategy used this idea more generally and show that simply copying and pasting random patches from one image to another and modifying the output probabilities improves both the classification and localization performance. Finally, PuzzleMix \cite{kim2020puzzle} explored CutMix combined with saliency information and image statistics to further improve the performance. In the augmentation method, we propose the way to incorporate human knowledge can be seen as a supervised version of CutMix. We use the object location to copy and replace the objects in the dataset in the images with mismatched attention map and true object location.

\section{Methods}
\subsection{Multi-Scale Attention Branch Network}
In \cite{fukui2019attention}, a dual attention and perception branch architecture is used to generate the attention map and prediction simultaneously. The output from the third convolutional block (which is referred to as the ``feature extractor'') is fed into the attention branch which produces the CAM output and the attention map. The attention map is then combined with the feature extractor output and fed into the perception branch which outputs the probability of each class. The CAM output from the attention branch and the probability output of the perception branch are trained on cross-entropy loss simultaneously. The training can then be done in an end-to-end manner in both branches. The loss can be written as,

\begin{equation}
\mathcal{L}(x_i) = \mathcal{L}_{attn}(x_i)+\mathcal{L}_{cls}(x_i) 
\label{eq:abnloss}
\end{equation}
where $\mathcal{L}_{attn}(x_i)$ and $\mathcal{L}_{cls}(x_i)$ are the cross-entropy losses for the attention and perception branch respectively for a training sample $x_i$.

\begin{figure}        
    \centering
    \includegraphics[width=1\columnwidth]{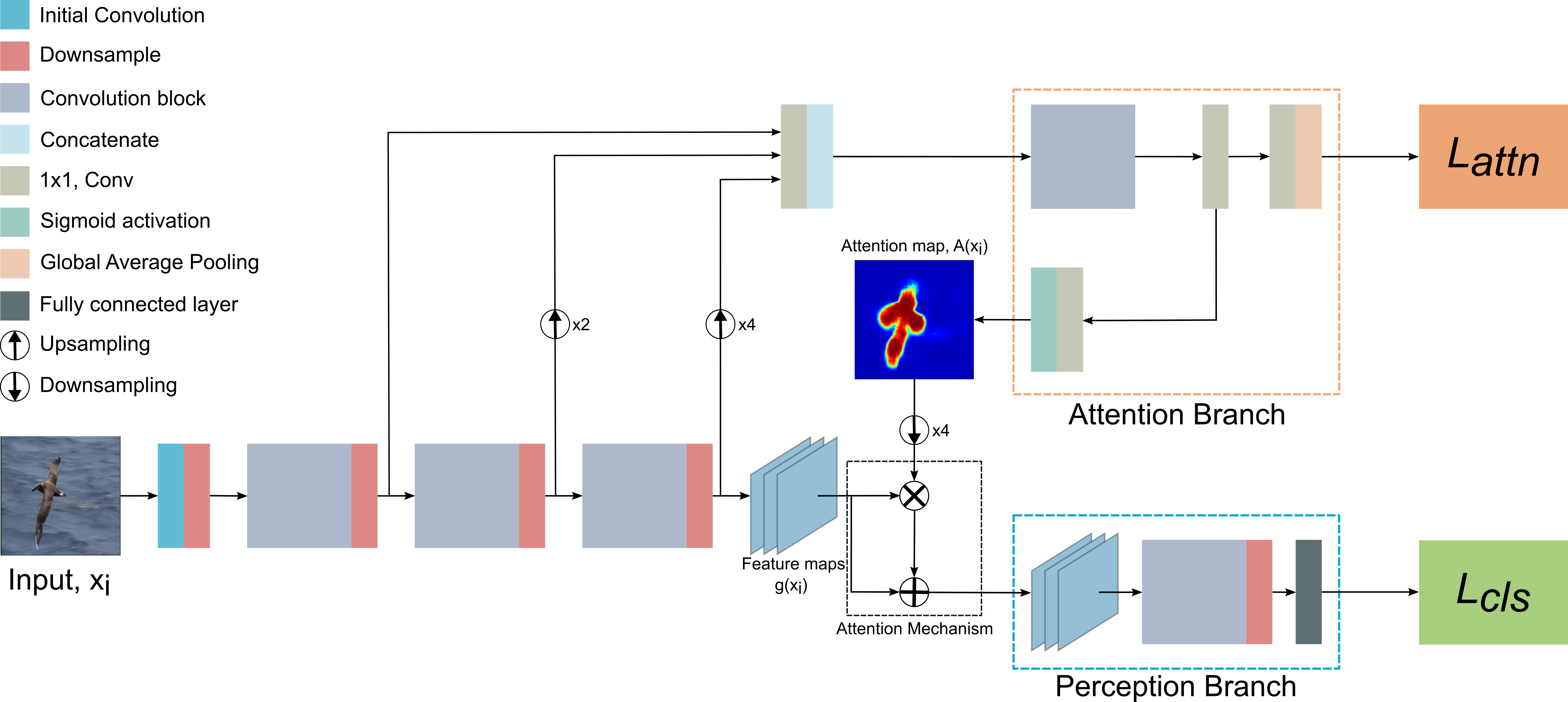}
    \caption{Overview of our proposed MSABN framework.}
    \label{fig:model}
\end{figure}

Unlike \cite{fukui2019attention}, where the attention branch is only fed the output of the third convolutional block, the attention branch of our multi-scale attention branch network (MSABN) is fed with input from the first and second convolutional blocks as well. Figure \ref{fig:model} shows the MSABN architecture. The output from the second and third convolutional blocks is upsampled to match the size of the first convolutional block. The output from each of the three convolutional blocks is then passed through $1\times1$ convolutional blocks such that the total number of channels after concatenating the outputs match the input channels of the attention branch.  Our suggested architecture helps the network accumulate hierarchical information at different scales to produce a fine-grained attention map and improve performance. The resolution of the attention map in MSABN is also higher than in ABN as a result of the upsampling.

\subsection{Puzzle Module to Improve Fine-grained Recognition}
\begin{figure}        
    \centering
    \includegraphics[width=0.9\columnwidth]{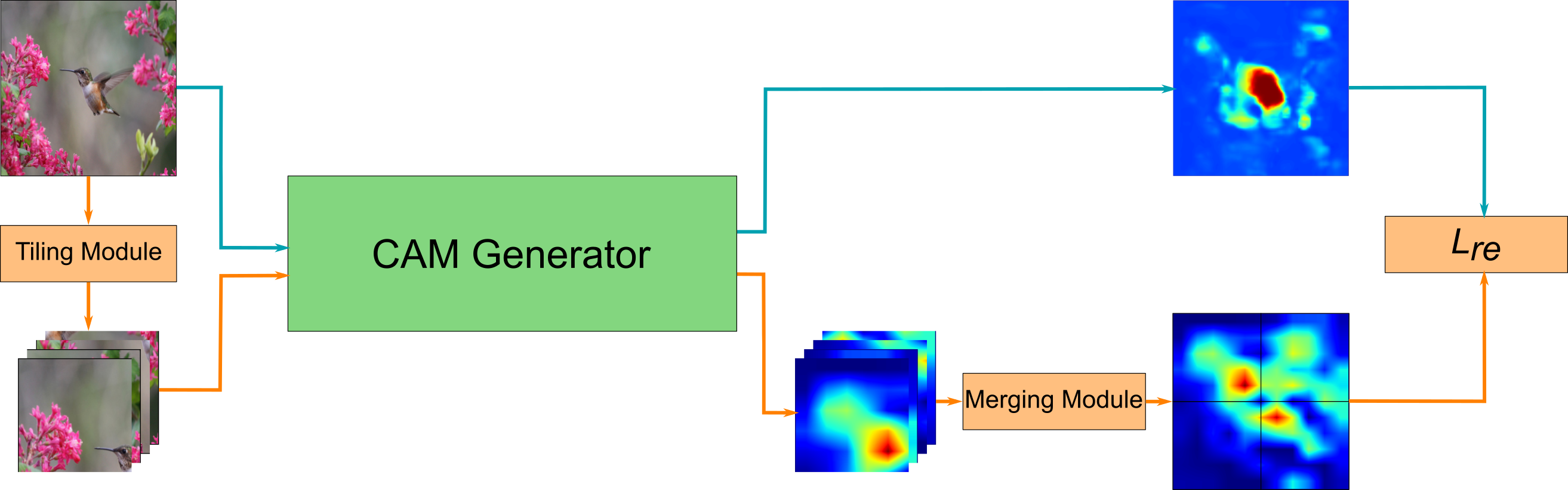}
    \caption{Overview of the Puzzle Module. The tiling module divides the image into non-overlapping tiles which are then fed into the CAM generator and the merging module merges the tiled CAMs that are then compared with the original CAM.}
    \label{fig:puzzle_fig}
\end{figure}

The puzzle module consists of a tiling and a merging module which aims at minimizing the difference between the merged features from tiled patches in an image and the features from the original image. The tiling module generates non-overlapping tiled patches of the input image. The patches are fed into the network and the merging module merges the patches' outputs into CAMs. The L1 loss between the CAM of the target class from the original image and the reconstruction is then calculated and used for training. Figure \ref{fig:puzzle_fig} illustrates the puzzle module. The overall loss for training MSABN models with the puzzle module becomes:

\begin{equation}
\mathcal{L}(x_i) = \mathcal{L}_{attn}(x_i)+\mathcal{L}_{cls}(x_i) +\mathcal{L}_{re}(x_i), 
\label{eq:abnpuzzleloss}
\end{equation}

\noindent where $\mathcal{L}_{re}(x_i)$ is the L1 loss between the original image CAM and the reconstructed CAM.

The puzzle module was introduced in \cite{jo2021puzzle} for weakly-supervised segmentation. It was shown to improve the segmentation performance on the PASCAL VOC 2012 dataset \cite{everingham2010pascal} with CAM-based ResNet models. Since the attention branch is trained similarly to CAM models, we decided to apply the puzzle module to the attention branch to observe the effects on attention maps and the performance of the models.

\subsection{Embedding Human Knowledge with Copy-Replace Augmentation}
\begin{figure}        
    \centering
    \includegraphics[width=0.7\columnwidth]{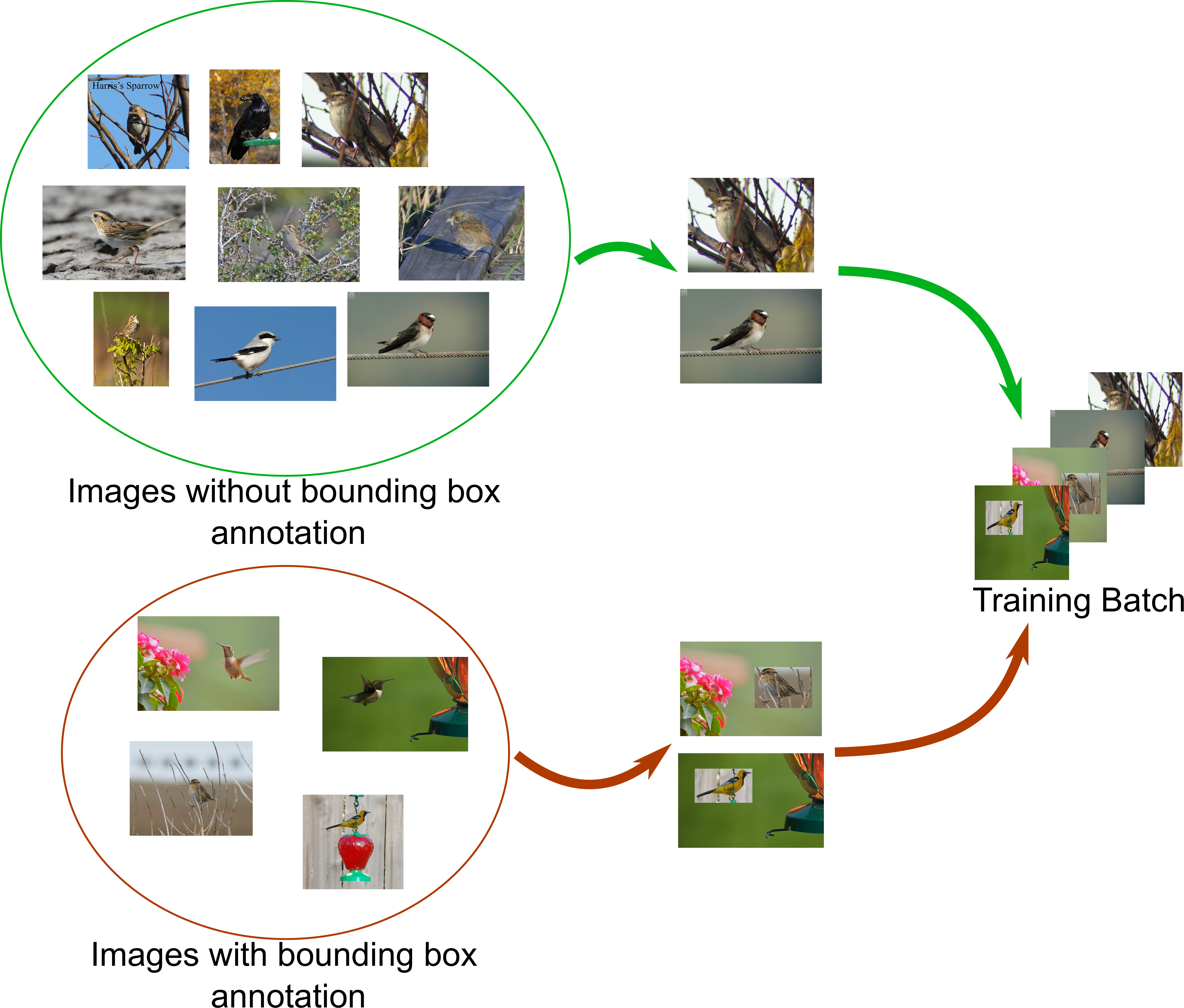}
    \caption{Overview of the copy-paste training scheme. In our case, objects with misaligned attention are annotated with a bounding box and, cut out, and replaced by other bounding box annotated objects.}
    \label{fig:bbox_fig}
\end{figure}

We describe a HITL pipeline with the proposed fine-tuning process incorporating human knowledge. First, the MSABN model is trained with images in the training set with corresponding labels. Then, the attention maps of the training samples are obtained from the best-performing model. In a real setting, the user can then inspect the attention maps and provide corrective input where the attention is mislocated in the form of a bounding box around the object. This is a softer form of user input compared to object boundary annotation and can be scaled to some degree without it becoming too cumbersome. Then, the training images are divided into two pools, one with bounding box annotations provided by the user and the other without. In the fine-tuning step, we train with all the training images and labels normally, \emph{but} we modify the images where the bounding box annotation was provided. For the images with provided annotations, the patch defined by the bounding box in an image is extracted, resized, and pasted onto the bounding box of another image from the same pool as illustrated in Figure \ref{fig:bbox_fig}. The resizing to the target bounding box size also acts as scaling augmentation during the training.

\section{Experiments}\label{sec:exps}
\subsection{Image Classification}

\subsubsection{Datasets:}
We evaluate MSABN for image classification on the CIFAR100 \cite{krizhevsky2009learning}, ImageNet \cite{deng2009imagenet} and DiagSet-10x \cite{koziarski2021diagset} datasets. CIFAR100 contains 60,000 images for training and 10,000 images for testing of 100 classes with an image size of 32$\times$32 pixels. ImageNet consists of 1,281,167 training images and 50,000 validation images of 1000 classes. Finally, the DiagSet-10x dataset is a histopathology dataset for prostate cancer detection with 256$\times$256 pixel images divided into 9 categories with different cancer grading, normal tissue, artifacts, and background. This dataset was included to test the performance of MSABN on texture data rather than object data. The dataset consists of 132,882 images for training, 25,294 for validation, and 30,086 for testing. In the experiments, the background class (empty, non-tissue regions of the slide) was excluded.

\subsubsection{Training details:}
For the CIFAR100 dataset, ResNet-20,-56,-110 \cite{he2016deep} models were used to evaluate the performance. The models were optimized with stochastic gradient descent (SGD) with a momentum of 0.9 and weight decay of 5e-4 for 300 epochs and a batch size of 256. The initial learning rate was set to 0.1 and was divided by 10 at 50$\%$ and 75$\%$ of the total number of epochs. Random cropping and horizontal flipping were used as augmentations, and model weights were initialized with Kaiming initialization \cite{he2015delving}. The experiments were repeated three times with different random seeds and the mean and standard deviation of the accuracy is reported (mainly to assert stability of the models). 

For the DiagSet-10x and ImageNet datasets, ResNet-101 and EfficientNet-B1 models were used. The models were optimized with SGD with a momentum of 0.9 and weight decay of 1e-4 for 90 epochs. A batch size of 512 was used for ResNet-101 experiments and 256 for EfficientNet-B1 experiments. The initial learning rate was set to 0.1 and was divided by 10 at 33$\%$ and 66$\%$ of the total number of epochs. For both datasets, random cropping of 224$\times$224 and horizontal flipping were used as augmentations. For DiagSet-10x vertical flipping and color jitter were also used. Due to a severe imbalance between classes in the DiagSet-10x dataset, the loss was weighted by the number of samples of each class. All model weights were initialized with Kaiming initialization. These experiments were only performed once due to computational constraints. EfficientNet-B1+ABN performance was not reported as the authors in \cite{fukui2019attention} didn't implement ABN on EfficientNets.

\subsubsection{Analysis of attention mechanism on CIFAR:} 
Following the analysis in \cite{fukui2019attention}, two different attention mechanisms (how to combine the attention map with the feature maps), namely $g(x)\cdot A(x)$ and $g(x)\cdot (1+A(x))$, were compared with the base (no attention branch) and ABN model which uses the $g(x)\cdot (1+A(x))$ mechanism. As can be seen from Table \ref{tab:cifar_perf}, both mechanisms outperform the ABN for all model versions. $g(x)\cdot (1+A(x))$ outperforms $g(x)\cdot A(x)$ slightly for the Resnet-20 and 110 models but performs slightly worse for ResNet-56 on average. The standard deviation for $g(x)\cdot A(x)$ mechanism is higher than for $g(x)\cdot (1+A(x))$ for all models. This might be due to the greater stability provided by the residual connection. We also noticed that the attention maps when using $g(x)\cdot A(x)$ have values close to 0.5 for the background as compared to $g(x)\cdot (1+A(x))$ where they were close to zero. Hence we decided to use $g(x)\cdot (1+A(x))$ mechanism as the default for this paper.

\begin{table}
\centering
\caption{Comparison of the accuracies (\%) on CIFAR100}
\begin{tabular}{c|c|c|c}
                    & ResNet-20      & ResNet-56      & ResNet-110     \\
\hline
BaseModel                           & 68.71±0.202 & 74.92±0.520 & 76.83±0.233 \\
BaseModel+ABN                          & 68.97±0.074 & 76.31±0.029 & 77.92±0.102 \\
BaseModel+MSABN ($g(x)\cdot A(x)$)     & 69.47±0.298 &	\textbf{77.16±0.419} & 78.40±0.254 \\
BaseModel+MSABN ($g(x)\cdot (1+A(x))$) & \textbf{70.06±0.025} &	76.78±0.149	& \textbf{78.68±0.121} \\
\hline
\end{tabular}
\label{tab:cifar_perf}
\end{table}

\subsubsection{Accuracy on ImageNet and DiagSet-10x:} 
\begin{figure}        
    \centering
    \includegraphics[width=0.9\columnwidth]{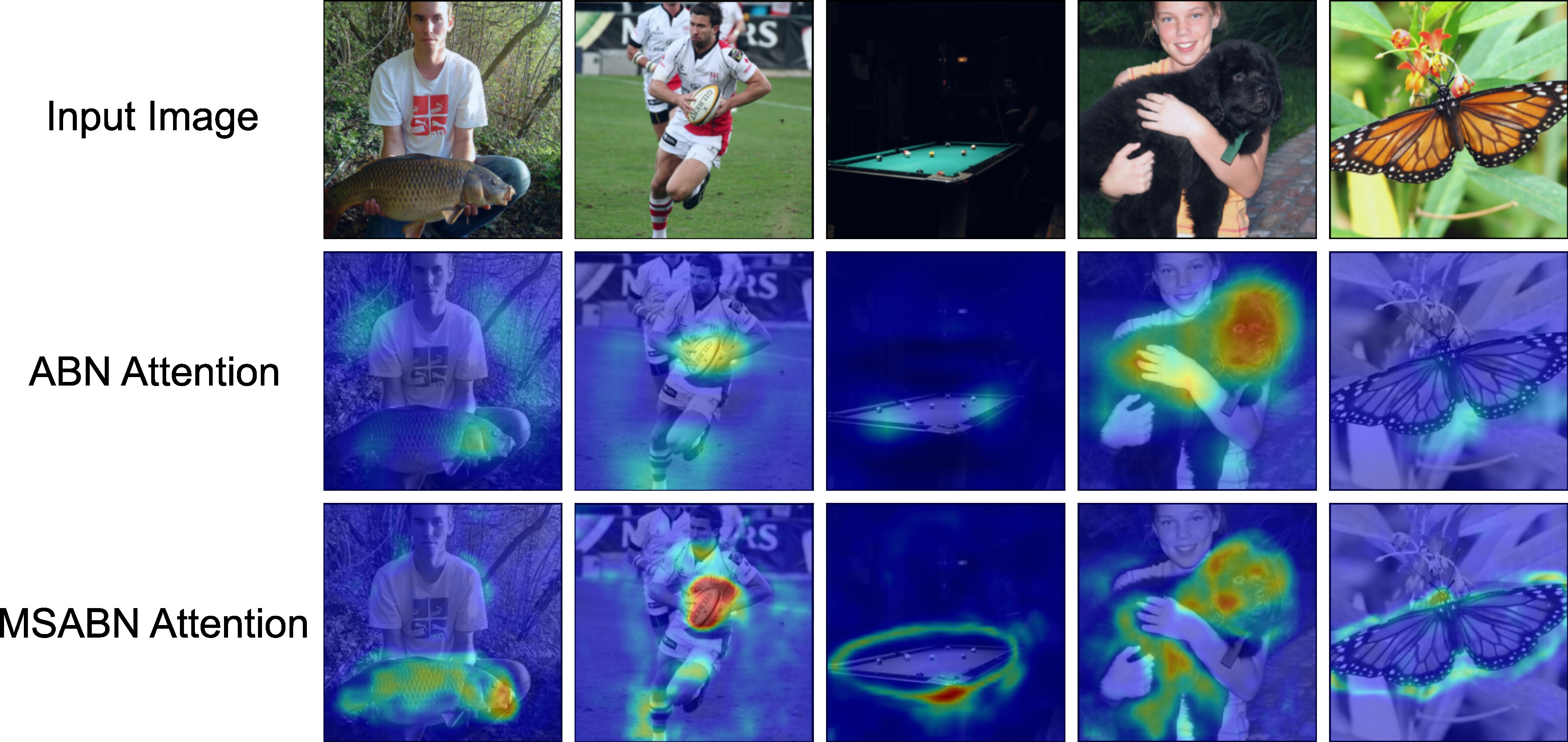} 
    \caption{Visualization of high attention regions in the images from the ImageNet dataset.}
    \label{fig:imagenet_results}
\end{figure}

Table \ref{tab:imagenet_perf} summarizes the results of the experiments on ImageNet. The MSABN model outperforms the ABN model by 0.35\% for ResNet-101 and the base model by 0.76\% and 1.41\% for ResNet-101 and EfficientNet-B1 respectively. The visual comparison of the attention in ABN and MSABN models is shown in Figure \ref{fig:imagenet_results}.
The MSABN attention has a better localization performance and highlights the object boundaries better than ABN attention.  

\begin{table}
\caption{Comparison of the accuracies (\%) on the ImageNet Dataset}
\centering
\begin{tabular}{c|c|c}
                     & ResNet-101 & EfficientNet-B1     \\
\hline
BaseModel              & 77.33 & 67.61\\
BaseModel+ABN          & 77.74 & -\\
BaseModel+MSABN        & \textbf{78.09} & \textbf{69.08}\\
\hline
\end{tabular}
\label{tab:imagenet_perf}
\end{table}

\begin{figure}        
    \centering
    \includegraphics[width=0.9\columnwidth]{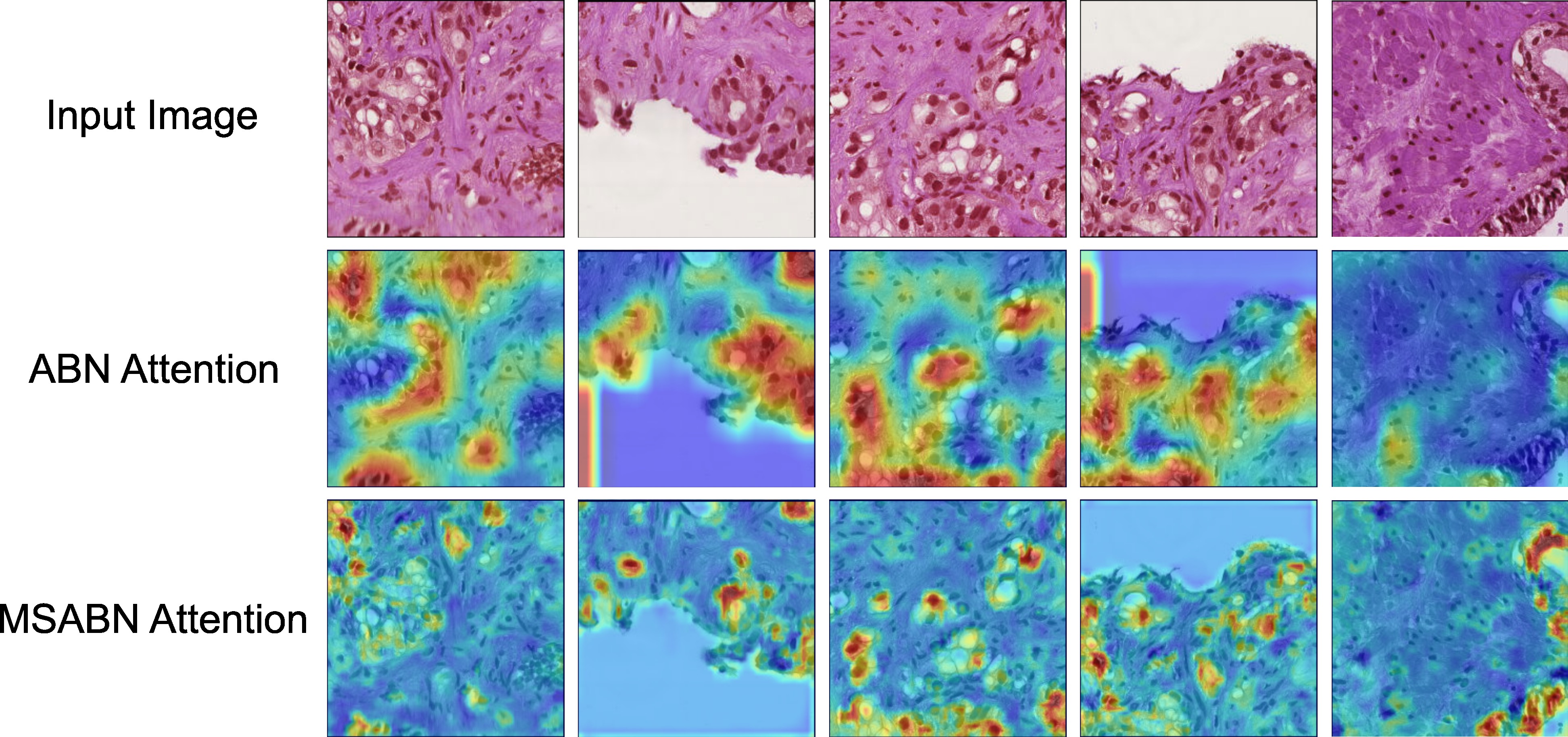}
    \caption{Visualization of high attention regions in the DiagSet-10x dataset.}
    \label{fig:diag_results}
\end{figure}

Table \ref{tab:diagset_perf} summarizes the results of the experiments on DiagSet-10x. In addition to accuracy, we also report the balanced accuracy to better estimate the performance on an unbalanced dataset. Here, the MSABN model outperforms the ABN model by 4.23\%/2.68\% for ResNet-101 and the base model by 2.34\%/2.72\% and 0.09\%/1.91\% for ResNet-101 and EfficientNet-B1 respectively. A visual comparison of the attention in ABN and MSABN models is shown in Figure \ref{fig:diag_results}. As compared to ABN, the MSABN attention highlights specific nuclei important for the grading. ABN attention maps also show an activation along the image boundary in the images with non-tissue regions present as an artifact. This artifact was not observed in the MSABN attention maps.

\begin{table}
\caption{Comparison of the accuracies (\%) on the DiagSet-10x dataset}
\centering
\begin{tabular}{c|c|c}
                     & \shortstack{ResNet-101 \\(Accuracy$/$Balanced Accuracy)} & \shortstack{EfficientNet-B1\\(Accuracy$/$Balanced Accuracy)}     \\
\hline
BaseModel              & 54.96 / 38.57 & 56.15 / 40.36\\
BaseModel+ABN          & 53.07 / 38.61 & -\\
BaseModel+MSABN        & \textbf{57.30 / 41.29} & \textbf{56.24 / 42.27}\\
\hline
\end{tabular}
\label{tab:diagset_perf}
\end{table}

\subsection{Fine-grained Recognition}

\subsubsection{Datasets:}
The performance of MSABN and Puzzle-MSABN was evaluated in two fine-grained recognition datasets namely CUB-200-2011 \cite{WahCUB_200_2011}, and Stanford Cars \cite{KrauseStarkDengFei-Fei_3DRR2013}. The CUB-200-2011 dataset contains 11788 images of 200 categories, 5994 for training and 5794 for testing. The Stanford Cars dataset contains 16185 images of 196 classes and is split into 8144 training images and 8041 testing images.

\subsubsection{Training details:}
The models used for evaluation of the fine-grained datasets were ResNet-50 \cite{he2016deep}, ResNext-50 (32$\times$4d) \cite{xie2017aggregated}, and EfficientNet-B3 \cite{tan2019efficientnet}. The models showcase different popular architectures and hence were chosen to benchmark the results. The models were optimized with SGD with a momentum of 0.9 and weight decay of 1e-4 for 300 epochs with a batch size of 16. The initial learning rate was set to 0.1 and was divided by 10 at 50$\%$ and 75$\%$ of the total number of epochs. The images were resized to 352$\times$352 pixels and the augmentations used were horizontal flip, color jitter, gaussian blur and noise, and solarize. All model weights were initialized with Kaiming initialisation. The experiments were run three times with different random seeds and the mean and standard deviation of the performance are reported.

Table \ref{tab:cub_perf} shows the accuracies of the models on the CUB-200-2011 dataset. MSABN improves upon the average performance of ABN models by 4\% and 2\% for ResNet-50 and ResNext-50 respectively. The performance gain compared with the base model is 14\%, 12\%, and 12\% for ResNet-50, ResNext-50, and EfficientNet-B3 respectively. Furthermore, the introduction of the puzzle module outperforms the ABN models by 6\% and 4\% respectively. The puzzle module outperforms the base models by 17\%, 14\%, and 13\% for ResNet-50, ResNext-50, and EfficientNet-B3, respectively. 

\begin{table}
\caption{Comparison of the accuracies (\%) on CUB-200-2011 Dataset}
\centering
\begin{tabular}{c|c|c|c}
                     & ResNet-50      & ResNext-50      & EfficientNet-B3     \\
\hline
BaseModel              & 40.62±1.985 & 44.68±0.816 & 53.41±0.792 \\
BaseModel+ABN          & 50.96±2.145 & 54.88±0.700 & - \\
BaseModel+MSABN        & 54.98±0.219 &	56.86±0.581 & 65.28±0.397 \\
BaseModel+MSABN+Puzzle & \textbf{57.31±0.645} & \textbf{58.87±0.592}	& \textbf{66.21±0.615} \\
\hline
\end{tabular}
\label{tab:cub_perf}
\end{table}

 Table \ref{tab:cars_perf}, shows the performance of the different models on the Stanford Cars dataset. The improvement with MSABN compared to the ABN model is not as significant here but slightly outperforms them on average. The performance gain compared with the base models however is 8\%, 3\%, and 2\% for ResNet-50, ResNext-50, and EfficientNet-B3, respectively. The introduction of the puzzle module outperforms ABN models by 3\% and 1\% respectively. The puzzle module outperforms the base models by 10\%, 4\%, and 3\% for ResNet-50, ResNext-50, and EfficientNet-B3, respectively. 

\begin{table}
\caption{Comparison of the accuracies (\%) on Stanford Cars Dataset}
\centering
\begin{tabular}{c|c|c|c}
                     & ResNet-50      & ResNext-50      & EfficientNet-B3     \\
\hline
BaseModel              & 78.70±1.062 & 84.00±0.817 & 86.86±0.495 \\
BaseModel+ABN          & 85.59±0.396 & 87.34±0.408 & - \\
BaseModel+MSABN        & 86.81±0.991 &	87.39±0.526 & 88.90±0.246 \\
BaseModel+MSABN+Puzzle & \textbf{88.25±1.492} & \textbf{88.32±1.184}	& \textbf{89.92±0.189} \\
\hline
\end{tabular}
\label{tab:cars_perf}
\end{table}

\begin{figure}        
    \centering
    \includegraphics[width=0.9\columnwidth]{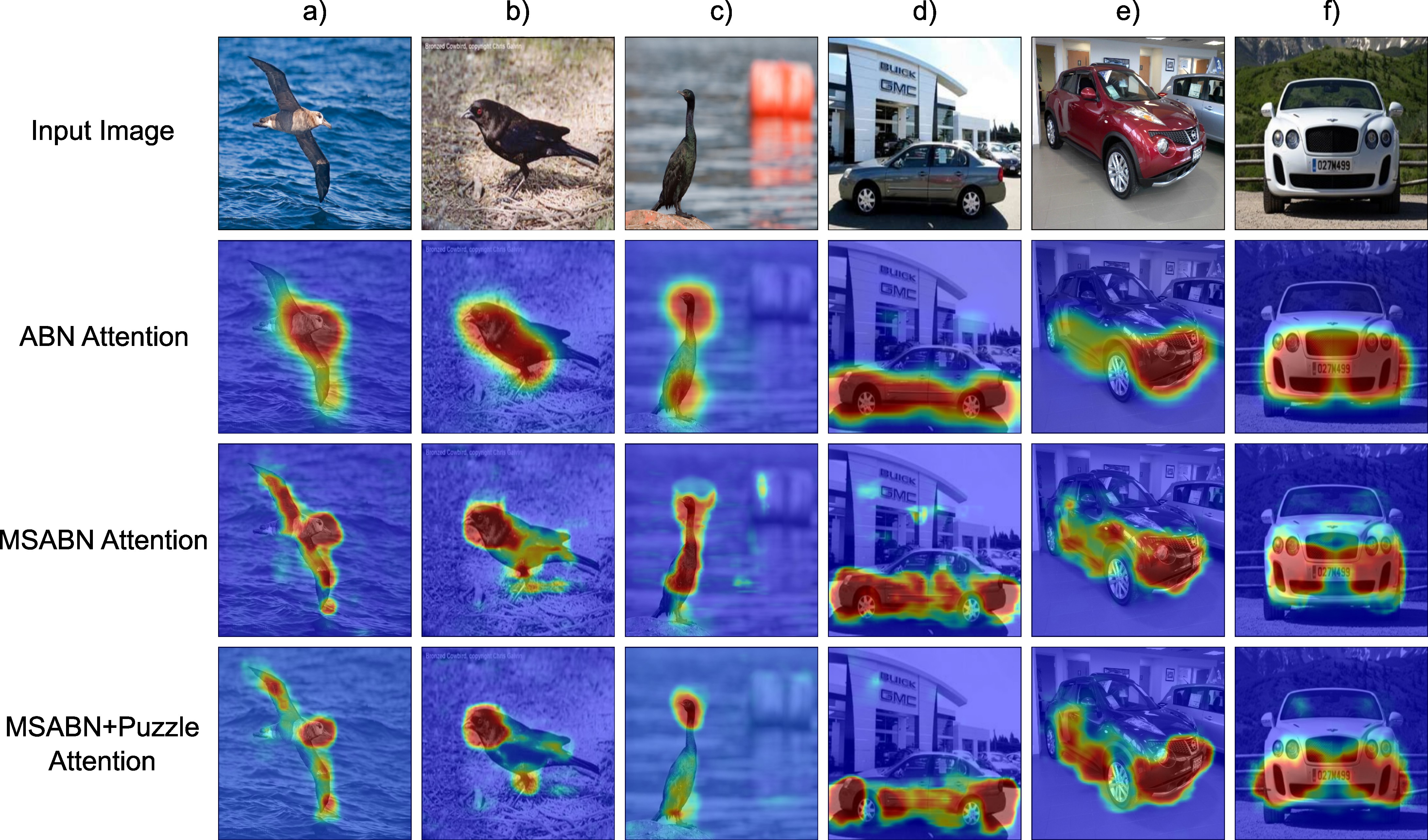} 
    \caption{Visualization of attention maps on the CUB-200-2011 (a, b, and c) and Stanford Cars (c, d, and e) dataset.}
    \label{fig:fine-grain_results}
\end{figure}

Visualization of the attention maps for CUB-200-2011 and Stanford Cars datasets are shown in Figure \ref{fig:fine-grain_results}. Compared to ABN attention maps, MSABN maps are able to delineate the object boundaries better and provide information about the discriminative regions in the image. For CUB-200-2011 dataset, most of the attention in MSABN and MSABN+Puzzle is focused on the key attributes like the bill, wings, or legs of the bird. The puzzle module performs significantly better than the rest of the model configurations here. This can be attributed to the effective regularization puzzle module provided in the case of small datasets and complex images (birds with different angles and actions). For Stanford Cars, the attention is focused on roughly similar regions depending on the car pose and form with different configurations. The performance improvement of MSABN and MSABN+Puzzle is not as significant here as for CUB-200-2011. This can be attributed to the nature of objects, i.e, cars, which have a well-defined shape as compared to birds which will have different shapes depending on the action like, flying, swimming, sitting, or standing.

\subsection{Attention Editing Performance}
\subsubsection{Datasets:}
We used the previously trained MSABN models on the CUB-200-2011 and Stanford Cars datasets to demonstrate the attention editing process. In both the datasets, bounding boxes of the objects within the image are available and are used here to mimic user input in the HITL pipeline in controlled experiments.

\subsubsection{Experimental setup:}
We measured the accuracy with respect to the number of annotated samples (bounding boxes) ``provided by the user'' to estimate the human effort required to achieve performance gain. To decide which samples to ``annotate'', the attention maps of the training data were saved and a binary map was created by thresholding the intensity values at 0.2. Next, the ratio of attention inside and outside the bounding box of the total attention in the binary image was calculated. The number of samples for which the fraction of attention outside $frac_{attn\_out}$ was above a threshold $\lambda_{out}$ were put in the bin of copy-replace augmentation for the fine-tuning training step. For the CUB-200-2011 dataset, the performance at $ \lambda_{out}\in\{0.0,0.1,0.2,0.3,0.4,0.5,0.6,0.7\}$ was obtained and for Stanford Cars, the performance was obtained at $\lambda_{out}\in\{0.0,0.1,0.2,0.3,0.4,0.5\}$. We stopped at 0.7 and 0.5 respectively because the number of training samples were too low (less than 30) in some configurations to observe meaningful changes above these thresholds. We chose this way instead of measuring performance at different fractions of the total number of training samples to better infer the trends of attention localization in the datasets and models. If the majority of the attention is focused inside the object box, the number of samples where $frac_{attn\_out} > \lambda_{out}$ would be much lower for higher thresholds. This method puts the images with the worst attention localization first which of course ``boosts'' the reported results. However, a case can be made that this mimics a potentially real scenario where a user corrects or edits discovered errors for fine-tuning a model: The training samples to show the user can be sorted either on the basis of wrong predictions first or the amount of overall attention (since the models might focus on the background) to be shown first, which will make user input a little less time-consuming. The results were compared with results from vanilla fine-tuning with all the training images to better compare the effects of the copy-replace augmentation.

\subsubsection{Training details:} We evaluated all the MSABN models used in the fine-grained recognition. The models were optimized with SGD with a momentum of 0.9 and weight decay of 1e-4 for 50 epochs with a batch size of 16. The initial learning rate was set to 0.1 and was divided by 10 at 50\% and 75\% of the total number of epochs. No augmentations were applied in the retraining to document the effects of only the copy-replace augmentation. The experiments were repeated three times for each model's (Resnet, ResNext, and EfficientNet) three repetitions in the previous section.

\begin{figure}        
    \centering
    \includegraphics[width=0.95\columnwidth]{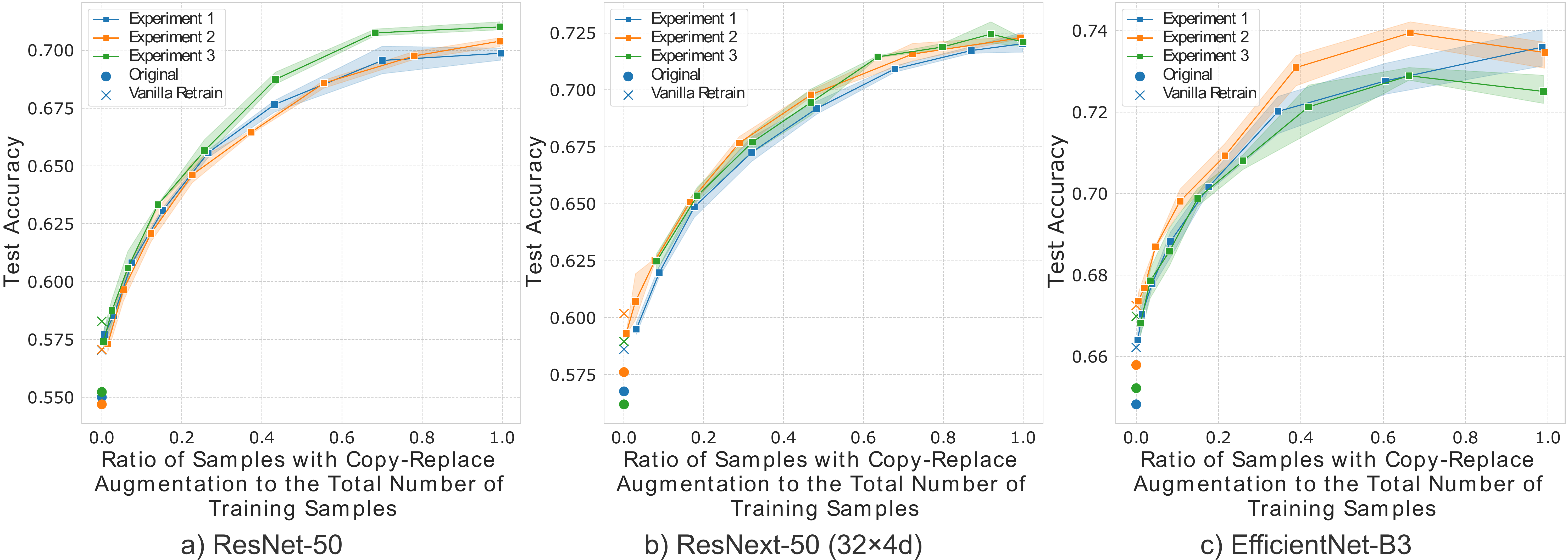} 
    \caption{Fine-tuning performance of different MSABN models on the CUB-200-2011 dataset with different ratios of training samples used for copy-replace augmentation.}
    \label{fig:cub_retrain}
\end{figure}

\begin{figure}        
    \centering
    \includegraphics[width=0.95\columnwidth]{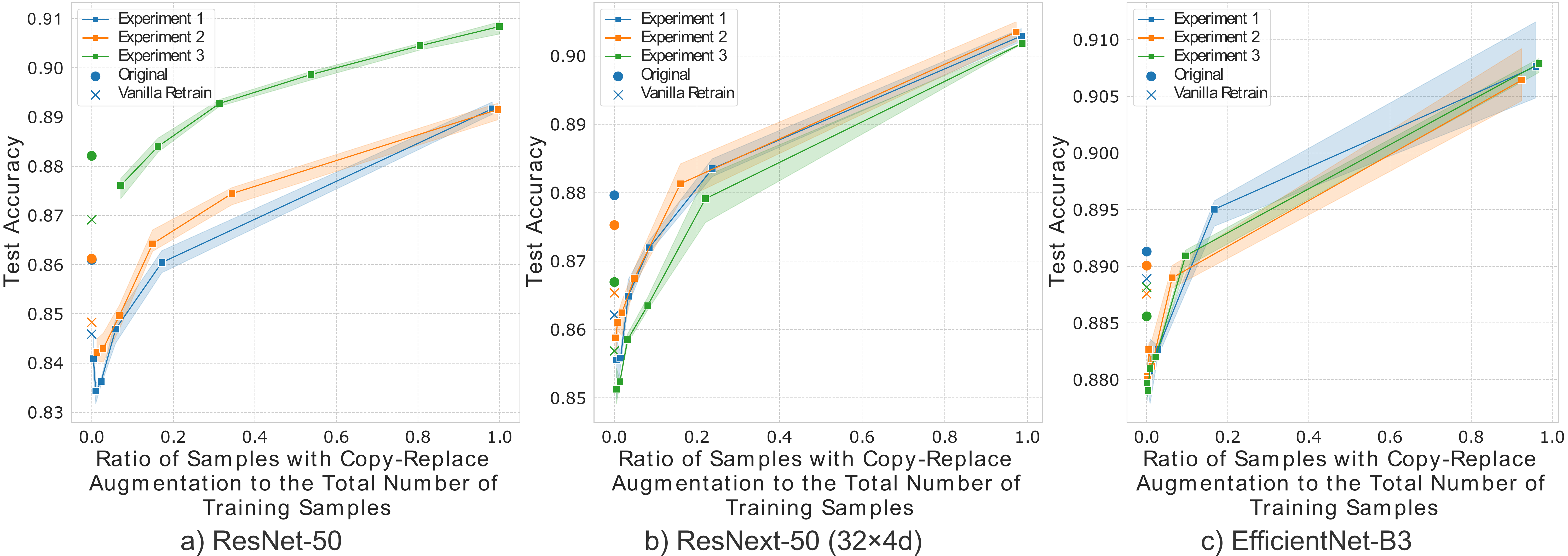} 
    \caption{Fine-tuning performance of different MSABN models on the Stanford Cars dataset with different ratios of training samples used for copy-replace augmentation.}
    \label{fig:cars_retrain}
\end{figure}

Figure \ref{fig:cub_retrain} and \ref{fig:cars_retrain} show the results of the fine-tuning with copy-replace augmentation in CUB-200-2011 and Stanford Cars datasets. As can be seen from the figures, fine-tuning improves the performance of the models. The accuracy gain compared to vanilla retraining with only 20\% of the training data augmented was approx. 6\%, 5\%, and  3\% for the MSABN models of ResNet-50, ResNext-50, and EfficientNet-B3 respectively for the CUB-200-2011 dataset. For Stanford Cars, the accuracy increased by 2\%, 1.5\% and 0.5\% respectively for the same amount of augmented training samples. The overall gain from augmenting almost all training samples compared with vanilla retraining was approx. 12.5\%, 12.5\%, and 6\% for MSABN models of ResNet-50, ResNext-50, and EfficientNet-B3, respectively for the CUB-200-2011 dataset. For the Stanford dataset, the overall gain was approx. 4.5\%, 3.5\%, and 1.5\%, respectively.

The distribution of data points along the x-axis in Figure \ref{fig:cub_retrain} is relatively uniform compared to Figure \ref{fig:cars_retrain} where most of the data points are located between zero and 0.2. This means that the number of training samples with $frac_{attn\_out}>\lambda_{out}$ changes proportionally with $\lambda_{out}$ for the CUB-200-2011 dataset but doesn't change much for the Stanford Cars dataset, which signals that the attention is more focused on the objects in the Stanford Cars dataset than in the CUB-200-2011 dataset. This can be attributed to the nature of objects in the datasets as mentioned earlier. 

The experiments showed that a significant performance gain can be achieved with limited and relatively crude human input. As mentioned earlier, the results are biased towards the worst examples, so there is a need to develop a smart sorting system that decides the order of images to be shown to the user.

We also noticed an interesting behaviour with vanilla retraining on the two datasets. While the accuracy after retraining in CUB-200-2011 increased, the accuracy decreased with the Stanford Cars dataset for all the models. The training loss curves (not shown) indicate that both models were destabilized (training loss increases) at the beginning of the training, however, in CUB-200-2011 dataset the models converged to a minima with lower validation loss than that of the initial training of the models. The observed effect in the Stanford Cars dataset was the opposite. Further experiments to determine the cause were not done, instead, the performance was compared to the vanilla retrain as the benchmark.

\textbf{Considerations:} It's worth noting that due to upscaling of the outputs of intermediate blocks, the attention branch now has to process 16x more input values. This increases the computational cost of the MSABN models. We noticed that it takes roughly twice the time to train MSABN models compared to ABN models. The puzzle module increases the cost further as every image is processed twice, once as original and other as tiled sub-images during training, however, the model behaves like the MSABN model during inference.

\section{Conclusion}
In this paper, we have presented a multi-scale attention branch network that greatly improves classification performance and also provides more accurate and detailed attention maps. We have evaluated the accuracy of MSABN for image recognition and fine-grained classification on multiple datasets and it was shown to outperform the ABN models. We also showed that using the puzzle module for fine-grained recognition increases the performance of the MSABN models. In addition, we introduced a HITL learning framework that inserts human knowledge in form of object bounding boxes and shows that this is an effective way of improving the performance further. The code is publicly available at \url{https://github.com/aktgpt/msabn}.

\textbf{Acknowledgments}. The work was supported by the Swedish Foundation for Strategic
Research (grant BD15-0008SB16-0046) and the European Research Council (grant ERC-2015-CoG 683810). The computations were enabled by the supercomputing resource Berzelius provided by National Supercomputer Centre at Linköping University and the Knut and Alice Wallenberg foundation.

\bibliographystyle{splncs04}
\bibliography{bib}
\end{document}